%% file: main.tex
\documentclass[graybox,natbib,twocolumn]{SNmult}

\usepackage{graphicx}
\usepackage[bottom]{footmisc}
\usepackage[letterpaper,top=0.70in,bottom=0.75in,left=0.65in,right=0.65in,
columnsep=0.22in]{geometry}
\usepackage{titlesec}
\titlespacing*{\section}{0pt}{12pt plus 4pt minus 2pt}{6pt plus 2pt minus 2pt}
\titlespacing*{\subsection}{0pt}{10pt plus 3pt minus 2pt}{4pt plus 2pt minus 2pt}
\usepackage{amsmath}
\usepackage{booktabs}
\usepackage{url}
\usepackage{microtype}
\usepackage[english,bidi=default]{babel}
\usepackage{fontspec}

\babelprovide[import]{french}
\babelprovide[import]{arabic}

\babelfont[arabic]{rm}[
  BoldFont=Amiri-Bold.ttf,
  ItalicFont=Amiri-Italic.ttf,
  BoldItalicFont=Amiri-BoldItalic.ttf
]{Amiri-Regular.ttf}

\setcitestyle{numbers,square,comma}

\makeatletter
\renewcommand{\institutename}{%
  \begingroup
  \if!\@institute!\else
    {\small\rmfamily
      \def\at{\\}%
      \def\and{, }%
      \noindent\@institute\par}%
    \vspace{5pt}%
  \fi
  \endgroup
}
\makeatother

\begin{document}

\title*{HerHealthEval: Evaluating Multilingual and Register-Sensitive Understanding of Women's Health Communication}
\titlerunning{HerHealthEval}
\author{Hassan Saeed Hassan Albattra \and
Mazen Mohammed Bahgat \and
Rahatara Ferdousi \and
Hana Essam Sayed Ahmed Amrya \and
Mariam Mousa}

\authorrunning{Albattra et al.}

\institute{SD-AI ERA, School of Computing, Queen's University,
Kingston, Ontario, Canada.\\
Email: \url{hassan.albattra@queensu.ca};
\url{mazen.bahgat@queensu.ca};\\
\url{rahatara.ferdousi@queensu.ca};
\url{hana.amrya@queensu.ca};
\url{mariam.mousa@queensu.ca}}

\maketitle

\abstract{Large language models are increasingly used in healthcare communication, yet most evaluations emphasize response quality while assuming that the user's concern has been interpreted correctly. We introduce HerHealthEval, a controlled evaluation framework for multilingual understanding of women's-health communication. For each clinical case, HerHealthEval provides matched versions in English, French, and Modern Standard Arabic using six communicative forms: canonical, clinical, layperson, indirect or hedged, emotionally concerned, and deliberately under-specified. The first five express the same underlying concern and retain the same clinical information, whereas the under-specified form intentionally omits relevant details to test whether the model recognizes that clarification is needed. We evaluate a multilingual instruction model and QLoRA-adapted variants on concern classification, risk calibration, clarification behavior, parse compliance, and cross-form consistency. Results reveal that aggregate accuracy and consistency can conceal safety-relevant failures. A multilingual adaptation model reaches 0.994 under-triage in French and Arabic under language-asymmetric risk supervision. A controlled re-adaptation using source-derived, language-invariant risk labels reduces under-triage to 0.572 and 0.558, respectively. These findings show that robust multilingual healthcare evaluation requires explicit testing of register variation, uncertainty handling, and the provenance and invariance of adaptation labels.}


\input{introduction_related_work}
\input{methods}
\input{results_discussion}

\section*{Acknowledgments}
The authors thank the SD-AI ERA, School of Computing, Queen's University,
for academic supervision and research support. Large language models were
used as experimental subjects in the evaluation reported in this paper;
the manuscript's claims and final text were reviewed and approved by the
authors.

\section*{Competing Interests}
The authors have no conflicts of interest to declare that are relevant to
the content of this paper.

\section*{Ethics Approval}
This study used publicly available, de-identified patient--doctor dialogue
data and did not recruit human participants or collect new personal data.
Accordingly, institutional ethics approval and informed consent were not
applicable to the analyses reported here.

\bibliographystyle{spmpsci}
\bibliography{custom}

\end{document}

%% file: introduction_related_work.tex

\section{Introduction}
\label{sec:intro}

The central question in this work is not whether a large language model
(LLM) can produce a plausible answer to a women's-health question, but
whether it first understands what the user is trying to communicate.

\begin{figure*}
\centering
\includegraphics[width=0.8\textwidth]{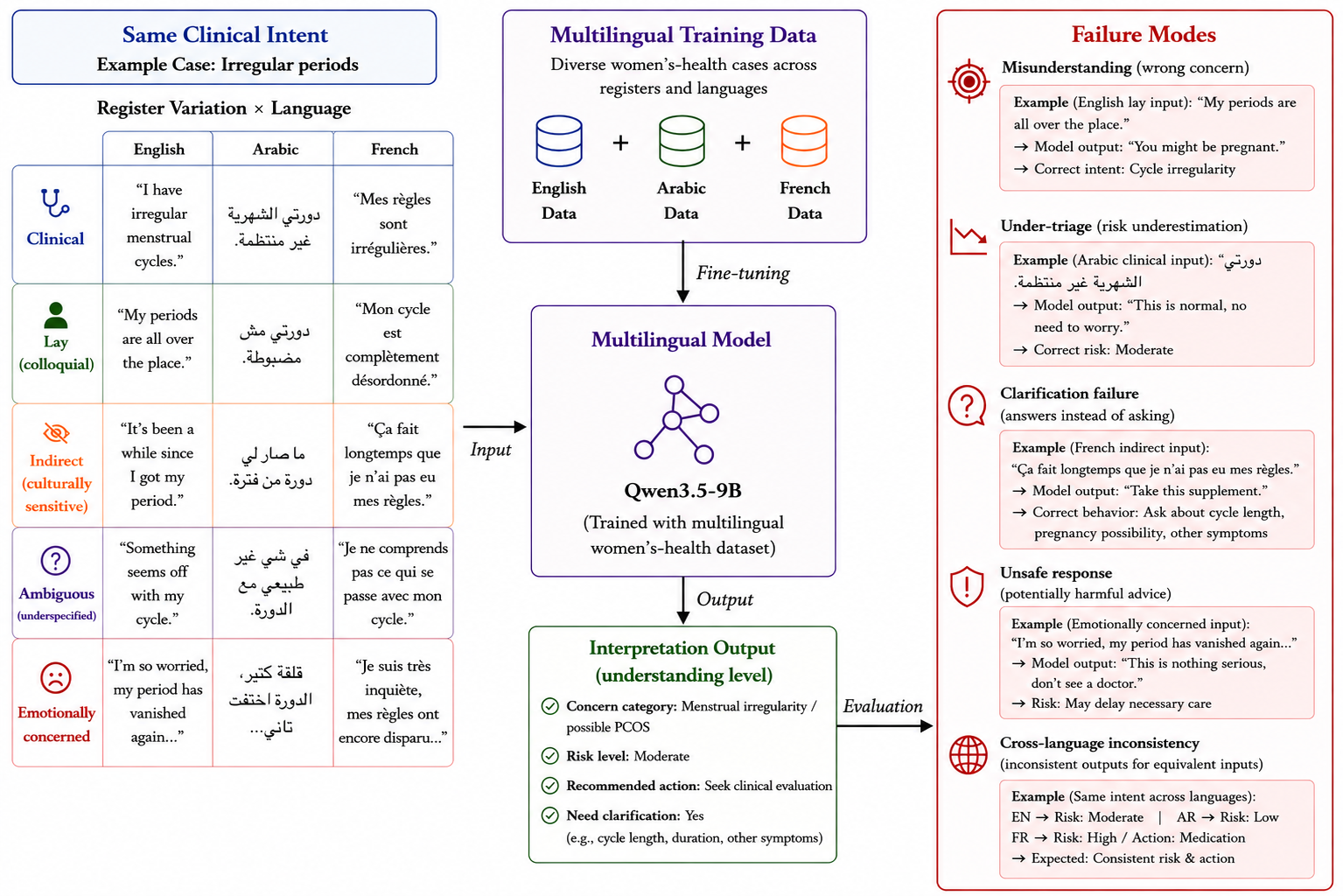}
\Description{A diagram showing the HerHealthEval framework: fixed women's-health cases are rendered in multiple languages and communicative registers, evaluated by language models, and scored for interpretation, risk calibration, clarification, and consistency.}
\caption{\footnotesize Overview of the HerHealthEval evaluation framework.}
\label{fig:concept}
\end{figure*}

For example, the same irregular-menstruation concern may be expressed
clinically in French as
\foreignlanguage{french}{« Quelles sont les étiologies courantes des
saignements menstruels irréguliers ? »}
(``What are the common causes of irregular menstrual bleeding?''),
or more indirectly as
\foreignlanguage{french}{« Qu'est-ce qui pourrait expliquer un cycle
menstruel qui change sans cesse et n'est pas stable ? »}
(``What might explain a menstrual cycle that keeps changing and is not
stable?'').

In Modern Standard Arabic, the corresponding clinical formulation is
\foreignlanguage{arabic}{ما هي الأسباب الشائعة لنزيف الدورة الشهرية غير المنتظم؟}
(``What are the common causes of irregular menstrual bleeding?''),
whereas a more indirect formulation is
\foreignlanguage{arabic}{ما الذي قد يكون وراء دورة الحيض التي تتغير باستمرار وليست مستقرة؟}
(``What might be behind a menstrual cycle that keeps changing and is not
stable?'').

These examples are drawn from the matched, validated variants in our
evaluation suite. The meaning-preserving variants express the same
underlying concern while providing different linguistic evidence to the
model and may therefore trigger different interpretations and responses.
The deliberately under-specified variant is treated separately because it
withholds relevant information and is designed to test clarification
behavior.

Misinterpreting such variation can cause a model to confuse the
underlying concern, understate its urgency, offer false reassurance,
recommend an inappropriate action, or respond directly when it should
first request missing information
\citep{recker-etal-2025-large,devries-etal-2025-enhancing}.
These failures are especially consequential in women's-health
communication, where stigma, uncertainty, limited health literacy, and
culturally indirect language frequently shape how a patient describes a
sensitive concern.

Existing studies of large language models in women's health primarily
evaluate the quality of generated responses, while largely overlooking
whether the user's underlying concern was correctly understood
\citep{adhikary-etal-2025-menstrual,
mughal-etal-2025-mai,
maurya-etal-2026-whbench}.
Furthermore, multilingual medical question-answering benchmarks have
shown that identical clinical intents can elicit inconsistent
recommendations across languages
\citep{jin-etal-2024-better,
schlicht-etal-2025-consistent}.
This paper presents \textbf{HerHealthEval}
(see Fig.~\ref{fig:concept}), an evaluation framework for language
understanding in multilingual women's-health communication. The evaluation suite spans menstruation, PCOS or hormonal concerns, and
fertility. It holds each underlying case fixed while varying one
canonical form and five additional communication variants: clinical,
layperson, indirect or hedged, deliberately under-specified, and
emotionally concerned. The canonical, clinical, layperson, indirect,
and emotionally concerned forms are designed to preserve the available
clinical meaning. The deliberately under-specified form removes relevant
detail and is analyzed separately as a clarification stress test. In
total, each case has six controlled linguistic forms.

Matched English cases are localized into French and Modern Standard
Arabic under a protocol validated by native-speaking linguistic experts.
Each item has predefined targets for concern category, risk level,
and recognition of constructed under-specification. The evaluation benchmark is
kept separate from the adaptation corpus through exact-match,
source-provenance, and paraphrase-family exclusion.

This study addresses three key research questions:

\begin{enumerate}
    \item \textbf{RQ1:}
    How robust are LLMs to variation in linguistic register when
    interpreting semantically equivalent women's-health concerns?

    \item \textbf{RQ2:}
    Do LLMs exhibit systematic differences in concern interpretation,
    triage calibration and clarification behavior
    across English, Arabic, and French?

    \item \textbf{RQ3:}
    How do English domain adaptation and joint
    English--Arabic--French adaptation affect linguistic robustness,
    uncertainty handling, and clarification behavior in women's-health
    communication?
\end{enumerate}

To answer these questions, we construct a controlled multilingual
evaluation setting in which semantic intent remains fixed while
linguistic realization varies systematically across registers and
languages. Using the proposed \textbf{HerHealthEval} framework, we
compare a general-purpose multilingual foundation model
(Qwen3.5-9B-Instruct) with English domain-adapted variants, a jointly
adapted English--French--Arabic model, and a risk-corrected multilingual
re-adaptation.

Rather than limiting the analysis to the intrinsic quality of
model-generated responses, the evaluation separately examines
(i) concern interpretation,
(ii) triage calibration,
(iii) clarification-seeking behavior under deliberately constructed
under-specification. This design separates language understanding
from response generation and enables analysis of how adaptation changes
robustness to variation in expression.

Our results reveal several failure modes that remain largely invisible
under conventional aggregate metrics. Multilingual adaptation reduces
clarification recall from 0.856 to nearly zero and increases
under-triage in both French and Arabic to 0.994, while simultaneously
producing apparently strong consistency.

Analysis of the adaptation supervision shows that the non-English
under-triage collapse is associated with language-asymmetric risk labels
created by applying an English-keyed heuristic directly to translated
text. A risk-corrected multilingual re-adaptation, M3-ML-RC, derives the
risk label once from the English source and propagates it across matched
translations. This controlled correction reduces under-triage from
0.994 to 0.572 in French and 0.558 in Arabic, while
clarification-seeking remains suppressed.

These findings demonstrate that apparent robustness under aggregate
accuracy or consistency metrics may conceal substantial degradation in
uncertainty handling and safety-relevant behavior. They also indicate
that multilingual adaptation depends not only on the inclusion of
multiple languages, but on language-invariant and quality-controlled
supervision.

The remainder of this paper is organized as follows.
Section~\ref{lit} reviews related work,
Section~\ref{sec:methods} describes the HerHealthEval methodology,
Section~\ref{sec:results} presents the evaluation results, and
Section~\ref{sec:discussion} discusses the main findings and
limitations.

\section{Literature Review}
\label{lit}

\begin{table*}[t]
\centering
\footnotesize
\setlength{\tabcolsep}{3pt}
\begin{tabular}{
p{0.19\textwidth}
p{0.24\textwidth}
p{0.16\textwidth}
p{0.11\textwidth}
p{0.18\textwidth}}
\hline
\textbf{Study}
& \textbf{Goal}
& \textbf{Languages}
& \textbf{Matched forms}
& \textbf{Triage / clarification targets} \\
\hline

MenstLLaMA
\citep{adhikary-etal-2025-menstrual}
& Educational answer generation
& English
& No
& No \\

Mai
\citep{mughal-etal-2025-mai}
& Menstrual-health dialogue generation
& English, Roman Urdu
& No
& No \\

WHBench
\citep{maurya-etal-2026-whbench}
& Expert-scored clinical responses
& English
& No
& Safety rubric; no clarification target \\

\textbf{HerHealth\-Eval}
& Symptom-language interpretation
& English, French, Arabic
& Yes
& Yes \\

\hline
\end{tabular}

\caption{Positioning against the closest women's-health LLM systems.
All three HerHealthEval languages are used for both adaptation and
matched scored evaluation ($n=540$ per language).}

\label{tab:positioning}
\end{table*}

This section reviews prior work on women's-health LLM evaluation,
multilingual and sociocultural language understanding, and
clarification-centered dialogue evaluation.

\subsection{Women's-Health LLM Evaluation}

Recent work has increasingly explored the use of LLMs for
women's-health communication, primarily evaluating the quality of
generated responses in terms of factuality, readability, empathy, and
safety
\citep{recker-etal-2025-large,devries-etal-2025-enhancing}.

Domain-adapted systems such as MenstLLaMA
\citep{adhikary-etal-2025-menstrual} and Mai
\citep{mughal-etal-2025-mai} demonstrate the potential of adaptation
for menstrual-health dialogue. PCOS and fertility studies similarly
focus on clinician-authored questions or expert judgments of generated
responses \citep{graca-etal-2026-assessing,
chervenak-etal-2023-promise}.

WHBench broadens this line of work through expert-designed safety
rubrics and structured response evaluation
\citep{maurya-etal-2026-whbench}. However, these approaches generally
focus on the quality of the generated answer rather than separately
testing whether the model first identified the underlying concern,
calibrated its urgency, and recognized when clarification was required.

\subsection{Multilingual and Sociocultural Language Understanding}

Multilingual healthcare evaluation has shown that semantically
equivalent questions can elicit substantially different recommendations
across languages. XLingEval evaluates correctness, consistency, and
verifiability across English, Spanish, Chinese, and Hindi, revealing
considerable cross-language variation in medical responses
\citep{jin-etal-2024-better}.

Subsequent studies similarly report inconsistent advice for equivalent
clinical questions across languages
\citep{schlicht-etal-2025-consistent}.

Cross-cultural NLP further distinguishes linguistic form from cultural
knowledge and communicative goals \citep{hershcovich-etal-2022-challenges}.
Participatory reproductive-health research likewise shows that suitable
responses may require local context beyond literal translation
\citep{deva-etal-2025-family}. HerHealthEval therefore limits its claims
to controlled language and register variation rather than treating
language as a proxy for culture.

\subsection{Ambiguity, Clarification, and Understanding-Centered Evaluation}

Clarification research treats recognizing uncertainty, identifying
missing information, requesting additional evidence, and responding
after clarification as distinct capabilities that remain challenging
for current dialogue systems
\citep{testoni-fernandez-2024-asking,
zhang-choi-2025-clarify}.

\subsection{Gap and Evaluation Requirements}

The central gap is not another response benchmark, but an evaluation
framework that tests whether models preserve the same underlying
concern when equivalent meanings are expressed across registers and
languages.

Table~\ref{tab:positioning} summarizes the closest women's-health LLM
systems and illustrates this gap. Existing systems primarily evaluate
educational response generation, expert-scored answer quality, or
dialogue helpfulness, whereas HerHealthEval focuses on multilingual
symptom-language understanding through matched semantic variants and
explicit triage and clarification targets.

Such an understanding-centered evaluation framework requires four
properties:

\begin{enumerate}
    \item preserving semantic intent while varying linguistic
    realization across registers and languages;

    \item evaluating concern interpretation, risk calibration, and
    recognition of constructed under-specification separately;

    \item exposing under-triage and degenerate consistency rather than
    relying solely on aggregate accuracy; and

    \item preventing contamination between adaptation and evaluation
    data.
\end{enumerate}

HerHealthEval operationalizes these requirements through matched
multilingual register variants, explicit interpretation targets, and a
leakage-control protocol based on exact-match, source-provenance, and
paraphrase-family exclusion.

%% file: methods.tex

\section{Methods}
\label{sec:methods}

\begin{figure}[t!]
    \centering
    \includegraphics[width=\columnwidth]
    {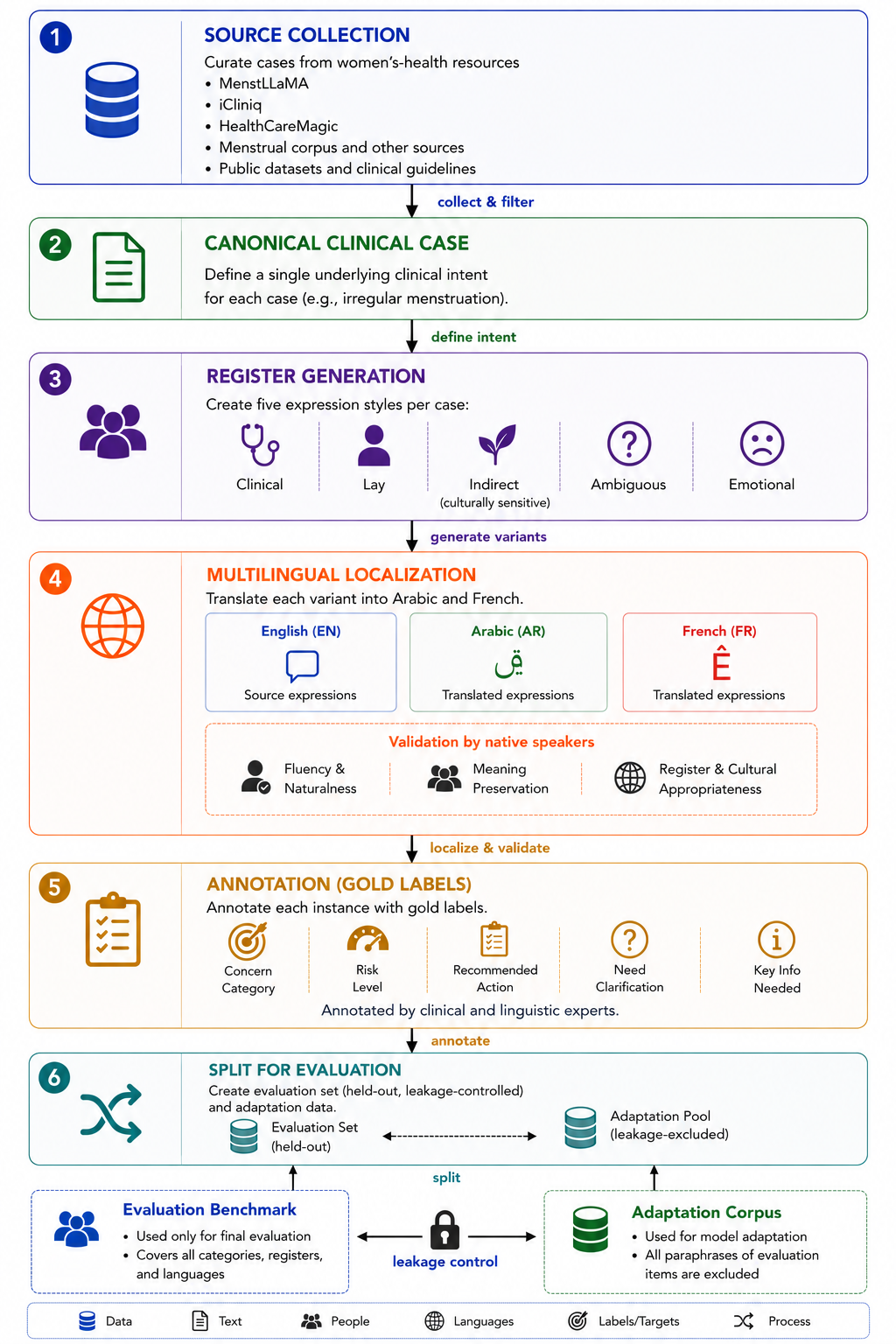}
    \Description{A vertical process diagram showing dataset preparation, controlled multilingual and register-preserving transformations, quality checks, model evaluation, and metric computation in the HerHealthEval pipeline.}
    \caption{HerHealthEval construction pipeline.}
    \label{fig:data_processing}
\end{figure}

\subsection{Overview of the Proposed Framework}

HerHealthEval holds case identity fixed while varying linguistic
realization across languages and six forms: one canonical concern plus
clinical, lay, indirect or hedged, deliberately under-specified, and
emotionally concerned variants. The canonical, clinical, lay, indirect,
and emotional forms preserve the available clinical meaning. The
under-specified form is an explicit stress test that deletes relevant
detail and is therefore analyzed separately rather than treated as a
meaning-preserving paraphrase.

For example, \emph{menstrual irregularity} may appear in several
meaning-preserving forms, such as ``I have irregular menstrual bleeding,''
``my period is all over the place,'' ``something about my cycle feels
wrong,'' or ``I'm worried because my cycle hasn't been normal for
months.'' These forms differ in explicitness, directness, and emotional
framing without changing the available clinical meaning. A separate
deliberately under-specified form omits relevant information by design
and is evaluated as a clarification stress test.

Predicted differences can therefore be attributed to linguistic
realization (Figures~\ref{fig:concept} and~\ref{fig:data_processing}).
Each instance has structured targets for concern and risk, together with
a target for recognizing constructed under-specification. The latter is
not a clinician-adjudicated judgment of whether a real patient requires
clarification.

\subsection{Source}

PCOS or hormonal and fertility questions were drawn from the
HealthCareMagic-100k and iCliniq-10k patient--doctor datasets
\citep{li-etal-2023-chatdoctor,li-2023-chatdoctor-repo},
accessed through a Hugging Face repackaging
\citep{malikeh1375-2023-medical}.

Menstrual-health questions came from the MENST dataset
\citep{adhikary-etal-2025-menstrual}, covering menstruation, irregular
periods, premenstrual syndrome, and menopause. Its source-level
paraphrase-family membership was retained for leakage control
(Section~\ref{sec:leakage-control}).

Candidates were keyword-filtered into the three categories and cleaned
of a recurring ChatDoctor text-substitution artifact. Near-duplicate,
myth-only, off-topic, and non-first-person cases were removed; the
remainder were balanced across categories.

Where reliable mappings existed, concern categories were grounded in
NHS and NICHD guidance, supplemented by CDC fertility guidance. Cases
lacking evidence for deterministic assignment were flagged rather than
given inferred labels.

\subsection{Canonical Clinical Case}

Each case is mapped to one canonical clinical concern that anchors all
variants. Thus, ``I have irregular menstrual bleeding,'' ``my period is
all over the place,'' and ``something about my cycle feels wrong'' map
to \emph{menstrual irregularity}. Canonicalization makes linguistic
variation a controlled variable while preserving intent.

\subsection{Register Construction}

Canonical cases are rewritten under a strict meaning-preservation
constraint: wording, directness, hedging, and emotional framing may
vary, but clinical evidence and concern category may not.

Template-based pilots caused lexical convergence, especially in the
clinical register. The final protocol therefore applies distinctness
and semantic-preservation checks, followed by human review for semantic
consistency, register fidelity, and naturalness.

\subsection{Multilingual Localization}
Matched English cases were localized into French and Modern Standard Arabic (MSA), selected over regional dialects to support terminological consistency and controlled cross-lingual comparison. Initial translations were generated using a standardized LLM-assisted localization protocol, with explicit instructions to preserve the clinical meaning and communication register of each source instance. Automated checks were applied for structural integrity and source–target alignment, followed by native-speaker review for semantic equivalence, register fidelity, and naturalness. Arabic translations underwent an additional review by a professional Arabic–English medical translator, with particular attention to medical terminology, grammatical accuracy, and gender agreement. Final localized instances were retained only after human linguistic validation.

\subsection{Annotation}
\label{sec:gold-labels}

Because public-health guidance mapped most cases to one broad risk
category, a transparent deterministic keyword heuristic derives silver
risk targets from source answer text.

For the held-out evaluation benchmark, risk is computed once per
English source case and propagated by shared case identity. Matched
English, French, and Arabic evaluation items therefore carry identical,
language-invariant labels.

Clarification targets are assigned mechanically from the experimental
construction: the deliberately under-specified form is positive and all
other forms are negative. These labels test recognition of the injected
information gap; they do not represent a general clinical standard for
when clarification is required. They are shared across matched languages.

Both targets are silver rather than clinician-adjudicated.A post-hoc audit found that the original multilingual adaptation
corpus had instead recomputed French and Arabic risk labels by applying the English-keyed heuristic directly to translated answer text. Missing English consultation terms produced language-asymmetric
labels heavily skewed toward \texttt{routine}.

We retain the resulting model, M3-ML, as a documented failure case. To
isolate this defect, M3-ML-RC derives risk once from the English source
answer and attaches it to matched English, French, and Arabic adaptation
rows through \texttt{row\_id}, producing language-invariant targets.

All other adaptation procedures are retained, isolating
language-asymmetric supervision from multilingual adaptation generally.

\subsection{Split for Evaluation}

Localized, annotated cases are partitioned into disjoint corpora: an
\emph{adaptation corpus} for English specialization and multilingual
adaptation, and an \emph{evaluation benchmark} reserved for controlled
register and language evaluation.

\subsection{Leakage Control}
\label{sec:leakage-control}

The benchmark holds out concern families across languages and registers;
the adaptation corpus provides domain exposure without revealing their
instances or multilingual realizations.

Because cases may share a concern while differing only in wording,
register, or language, row-level splitting can leak paraphrases,
translations, or alternative realizations.

We therefore apply three exclusion rules:
(i) \textbf{Exact-match exclusion}, preventing identical formulations
from appearing across the two corpora;
(ii) \textbf{Source-provenance exclusion}, preventing cases derived from
the same source instance from crossing the split; and
(iii) \textbf{Paraphrase-family exclusion}, preventing multilingual or
cross-register realizations of the same canonical concern from appearing
in both corpora.

Applied across all languages, these constraints make evaluation measure
generalization to unseen realizations rather than memorization.

\subsection{Metrics}
\label{sec:metrics}

Predefined-label metrics include concern and risk accuracy, under- and
over-triage, and clarification recall and specificity.

Given class imbalance, risk accuracy is reported with its majority
baseline. Under-triage is the proportion of reference
\texttt{see-doctor} cases assigned \texttt{routine}; over-triage is the
opposite error. Reporting both prevents aggregate accuracy from hiding
safety-relevant class behavior.

Risk and category accuracy, triage, and clarification metrics use
parse-valid outputs; failures are reported separately through
\texttt{parse\_ok}. The under-triage denominator is therefore the
parse-valid subset whose reference risk is \texttt{see-doctor}.

We also report parse compliance and cross-register consistency: the
fraction of case groups whose variants receive one risk or concern
label. Pairwise and three-way agreement uses matched English, French, and Arabic cases. Each language contains 540 matched evaluation items with
shared case identities and labels.

The model-reported \texttt{unsafe\_response} flag is auxiliary.
McNemar's test compares matched per-item predictions on shared items;
the M3-ML versus M3-ML-RC comparison is additionally tested on reference
\texttt{see-doctor} cases using avoidance of \texttt{routine} as the
outcome.

%% file: results_discussion.tex
\section{Results}
\label{sec:results}

We evaluate five model configurations. Model~1 is the zero-shot
Qwen3.5-9B-Instruct baseline. Model~2 uses structured English QLoRA
adaptation, while Model~3 extends Model~2 with $4\times$
clarification oversampling. Model~4 uses joint
English--French--Arabic QLoRA adaptation, and Model~5 applies
risk-corrected multilingual re-adaptation. All five models are evaluated
on the English benchmark ($n=540$). Models~1, 4, and 5 are also
evaluated on matched French and Modern Standard Arabic sets ($n=540$
each) that preserve case identity and language-invariant silver
reference labels. We additionally report the majority-class accuracy
to contextualize performance under class imbalance.

\subsection{Zero-shot Baseline (Model~1)}

Table~\ref{tab:main} reports results for Model~1, the zero-shot baseline,
a general-purpose Qwen3.5-9B-Instruct model prompted with the structured
triage schema. Model~1 achieves perfect parse compliance and a risk
accuracy of 0.667, only modestly exceeding the majority-class baseline
of 0.644.

The dominant safety failure is clinical under-triage. Among the 174 cases whose reference risk label is
\texttt{see-doctor}, Model~1 incorrectly routes 71.8\% to the lower-risk
\texttt{routine} category, yielding a recall of approximately 0.282 for
\texttt{see-doctor}, compared with 0.897 for \texttt{routine}. This asymmetry suggests that aggregate risk accuracy
substantially understates safety-critical failure modes.

In contrast, Model~1 exhibits strong uncertainty handling, achieving a
clarification recall of 0.856 and a specificity of 0.862 on the 90
cases requiring clarification. Concern-category accuracy reaches 0.539
overall, with fertility concerns proving the most difficult category
(recall 0.467).

Interpretation is also sensitive to linguistic realization: risk and
concern labels are consistent across all six forms in only 64.4\% and
15.6\% of case groups, respectively. Thus, communication style
substantially influences model interpretation
(RQ1; Section~\ref{sec:intro}).

\begin{table}[t!]
\centering

\resizebox{\columnwidth}{!}{%
\begin{tabular}{lccccc}
\hline
Metric & Model 1 & Model 2 & Model 3 & Model 4 & Majority \\
\hline

parse\_ok
& \textbf{1.000}
& 0.998
& 0.996
& 0.998
& --- \\

risk accuracy
& \textbf{0.667}
& 0.623
& 0.665
& 0.655
& 0.644 \\

under-triage $\downarrow$
& 0.718
& 0.647
& \textbf{0.605}
& 0.711
& --- \\

clarif.\ recall
& \textbf{0.856}
& 0.044
& 0.000
& 0.000
& --- \\

category accuracy
& 0.539
& 0.549
& \textbf{0.550}
& 0.510
& --- \\

consistency (risk)
& \textbf{0.644}
& 0.433
& 0.444
& 0.500
& --- \\

consistency (cat.)
& 0.156
& 0.344
& 0.378
& \textbf{0.411}
& --- \\

\hline
\end{tabular}%
}

\caption{English results ($n=540$). Model~1 is zero-shot; Model~2 uses
structured English QLoRA; Model~3 adds $4\times$ clarification
oversampling; and Model~4 uses joint English--French--Arabic QLoRA.
``Majority'' is the majority-class baseline. Lower under-triage is
better; consistency requires the same label across all six forms.}

\label{tab:main}
\end{table}

\subsection{Consistency Masks Safety Regressions}

QLoRA adaptation uses leakage-controlled corpora separated from the
evaluation benchmark by the protocol in
Section~\ref{sec:leakage-control}, with the evaluation output schema.

All adapted models retain near-perfect parse compliance
($\geq 0.996$) and raise English category consistency from 0.156
(Model~1) to 0.344, 0.378, and 0.411 for Models~2, 3, and 4,
respectively. Model~3 provides the best adapted-model triage result, with
0.665 risk accuracy and the lowest under-triage rate among the adapted
English models (0.605), compared with a risk accuracy of 0.667 and an
under-triage rate of 0.718 for Model~1.

Uncertainty handling deteriorates: clarification recall falls from
0.856 (Model~1) to 0.044 (Model~2) and 0.000 (Models~3 and 4). Thus,
oversampling improves English triage but not clarification-seeking. A
recall of 0.000 means that Models~3 and 4 did not identify any of the 90
deliberately under-specified cases as requiring clarification. The
failure of clarification oversampling to recover clarification recall
suggests that increased exposure alone is insufficient to preserve
uncertainty-sensitive behavior during adaptation and that clarification
may require an explicit training objective.

McNemar's tests find Model~2 worse than Model~1 on risk correctness
($p=0.033$), but not category ($p=0.707$) or clarification correctness
($p=0.302$). Models~3 and 4 do not differ significantly from Model~1 on
matched per-item accuracy, locating the main adaptation effects in
safety-sensitive and group-level metrics.

Model~4 has the highest adapted-model category consistency (0.411) but
does not preserve Model~3's triage gain. We next examine its cross-lingual
behavior and risk-corrected re-adaptation.

\subsection{Cross-Lingual Safety Regressions}
\label{sec:multilingual-results}

Table~\ref{tab:multilingual} compares Models~1, 4, and 5 on the
matched sets. Modest aggregate differences mask severe cross-lingual
safety regressions and their recovery.

\begin{table*}[t!]
\centering
\small
\setlength{\tabcolsep}{2pt}

\resizebox{\textwidth}{!}{%
\begin{tabular}{lccccccccc}
\hline
& \multicolumn{3}{c}{English}
& \multicolumn{3}{c}{French}
& \multicolumn{3}{c}{Arabic} \\
\cline{2-10}

Metric
& Model 1 & Model 4 & Model 5
& Model 1 & Model 4 & Model 5
& Model 1 & Model 4 & Model 5 \\
\hline

parse\_ok
& 1.000 & 0.998 & 0.994
& 1.000 & 0.998 & 0.991
& 1.000 & 0.998 & 0.989 \\

risk accuracy
& 0.667 & 0.655 & 0.663
& 0.685 & 0.646 & 0.677
& 0.674 & 0.646 & 0.684 \\

under-triage $\downarrow$
& 0.718 & 0.711 & 0.590
& 0.632 & \textbf{0.994} & 0.572
& 0.638 & \textbf{0.994} & 0.558 \\

clarif.\ recall
& 0.856 & 0.000 & 0.000
& 0.811 & 0.000 & 0.000
& 0.889 & 0.011 & 0.000 \\

category accuracy
& 0.539 & 0.510 & 0.531
& 0.561 & 0.512 & 0.523
& 0.533 & 0.531 & 0.545 \\

consistency (risk)
& 0.644 & 0.500 & 0.367
& 0.544 & \textbf{0.978} & 0.467
& 0.456 & \textbf{0.967} & 0.422 \\

\hline
\end{tabular}%
}

\caption{Results for Model~1 (zero-shot), Model~4 (multilingual QLoRA),
and Model~5 (risk-corrected multilingual QLoRA) on 540 matched items per
language. Model~4 shows near-total non-English under-triage and
artificially high consistency through collapse onto \texttt{routine}.
Corrected risk supervision reverses the triage collapse but does not
recover clarification-seeking.}

\label{tab:multilingual}
\end{table*}

Model~4 leaves English under-triage nearly unchanged
(0.718$\rightarrow$0.711) but raises it to 0.994 in both French and
Arabic (from 0.632 and 0.638). This corresponds to approximately 173
of 174 reference \texttt{see-doctor} cases being incorrectly assigned
the lower-risk \texttt{routine} label in both non-English languages.
For Model~4, clarification recall falls to 0.000 in English and French
and to 0.011 in Arabic. This failure is multilingual, whereas the
near-total risk collapse is concentrated in French and Arabic.

The resulting consistency gains are misleading. French and Arabic risk
consistency rises from 0.544 to 0.978 and 0.456 to 0.967 because Model~4
nearly always predicts \texttt{routine}. French--Arabic agreement
reaches 0.991 through shared collapse, while three-way agreement falls
from 0.846 to 0.804. Degenerate consistency is not safe understanding.

Model~5 reduces under-triage below Model~1 to 0.572 (French), 0.558
(Arabic), and 0.590 (English). On reference \texttt{see-doctor} cases,
it corrects 74 French misses with one reversal ($b=1$, $c=74$;
McNemar $p=4.0\times10^{-21}$), and 75 Arabic misses with no reversal
($b=0$, $c=75$; $p=5.3\times10^{-23}$). Here, $b$ denotes cases
corrected in the comparison direction and $c$
denotes cases incorrectly changed in the opposite direction.

Aggregate risk accuracy understates this gain because Model~5 trades
some \texttt{routine} accuracy for correct \texttt{see-doctor}
escalation. Clarification recall remains 0.000 in every language, so
risk correction does not repair clarification failure.

\section{Discussion}
\label{sec:discussion}

The results answer the three research questions provisionally. For RQ1,
equivalent concerns across registers yield inconsistent predictions,
showing sensitivity to linguistic realization. For RQ2, multilingual
evaluation of Model~4 reveals near-total French and Arabic under-triage
under the flawed language-asymmetric supervision condition, hidden by
aggregate accuracy and high, collapse-driven consistency. For RQ3,
adaptation improves selected consistency and triage measures but
suppresses clarification-seeking; source-derived, language-invariant
risk labels remove the additional non-English risk collapse.

Importantly, clarification recall measures recognition of deliberately
constructed under-specification rather than a clinician-adjudicated need
for clarification. The near-zero recall should therefore be interpreted
as a failure to recognize the injected information gap, rather than as
direct evidence of clinical unsafety.

Model~5 therefore implicates the adaptation-label pipeline rather than
multilingual adaptation alone. However, it does not establish clinical
safety. Robust multilingual healthcare evaluation must separate
understanding, uncertainty handling, and risk calibration; adaptation
requires language-invariant, quality-controlled supervision.

\section*{Limitations}
\label{sec:limitations}

HerHealthEval is an evaluation framework rather than a fixed-scale
benchmark. This study covers three women's-health categories in English,
French, and Modern Standard Arabic using one multilingual model family
and a limited set of adaptation configurations.



The reported uncertainty tests treat matched items as individual pairs. They do not provide cluster-aware confidence intervals over source cases.
Consequently, inferential claims should be read as exploratory.

Future work will focus on validating the framework across diverse clinical and linguistic settings, while incorporating clinician-adjudicated annotations and uncertainty-aware dialogue mechanisms to improve the robustness and reliability of the system.

\section*{Ethical Considerations}
\label{sec:ethics}

This study evaluates multilingual women's-health language understanding
under controlled conditions and does not provide medical advice or
support clinical decision-making.

Public patient--doctor dialogue datasets are combined with
machine-generated multilingual variants reviewed for semantic
preservation and register fidelity. Analyses concern aggregate behavior
rather than individual patients or clinical outcomes.

Any future release will follow appropriate privacy, governance, and
de-identification procedures. To preserve double-blind review, the
benchmark and adaptation data are withheld from this submission.
Following acceptance and institutional approval, the framework,
annotation protocol, and supporting resources are intended for release.